# Comparative Study on Semi-supervised Learning Applied for Anomaly Detection in Hydraulic Condition Monitoring System

Yongqi Dong [#, *]
*Department of Transport and Planning*
*Delft University of Technology*
Delft, the Netherlands
Y.Dong-4@tudelft.nl

Kejia Chen [#]
*Department of Software Engineering*
*Zhejiang University*
Hangzhou, China.
chenkejia@zju.edu.cn

Zhiyuan Ma
*Department of Information and Electrical Engineering*
*Shanghai Normal University*
Shanghai, China
raphealma@163.com

# These authors contributed equally to this work and should be considered as co-first authors. * Corresponding author (Y.Dong-4@tudelft.nl).

***Abstract*—Condition-based maintenance is becoming increasingly important in hydraulic systems. However, anomaly detection for these systems remains challenging, especially since that anomalous data is scarce and labeling such data is tedious and even dangerous. Therefore, it is advisable to make use of unsupervised or semi-supervised methods, especially for semi-supervised learning which utilizes unsupervised learning as a feature extraction mechanism to aid the supervised part when only a small number of labels are available. This study systematically compares semi-supervised learning methods applied for anomaly detection in hydraulic condition monitoring systems. Firstly, thorough data analysis and feature learning were carried out to understand the open-sourced hydraulic condition monitoring dataset. Then, various methods were implemented and evaluated including traditional stand-alone semi-supervised learning models (e.g., one-class SVM, Robust Covariance), ensemble models (e.g., Isolation Forest), and deep neural network based models (e.g., autoencoder, Hierarchical Extreme Learning Machine (HELM)). Typically, this study customized and implemented an extreme learning machine based semi-supervised HELM model and verified its superiority over other semi-supervised methods. Extensive experiments show that the customized HELM model obtained state-of-the-art performance with the highest accuracy (99.5%), the lowest false positive rate (0.015), and the best F1-score (0.985) beating other semi-supervised methods.**

## I. INTRODUCTION

Hydraulic systems are applied in a wide variety of industries since they take advantage of the properties of fluids that can be delivered to individual components without loss of applied pressure. Compared to mechanical and electrical transmissions, hydraulic systems rely on the internal pressure of the fluid for delivering power. However, if fluids in hydraulic systems leak as a spray, they can cause serious problems [1]. Thus, with increasing industrialization and technological advancements, together with the demand for high-quality life, condition monitoring of hydraulic systems has become more and more significant to ensure their reliability and stability, especially for anomaly detection and prevention.

To monitor the operational status of hydraulic systems effectively, various controllable sensors are installed to collect data and establish monitoring systems to safeguard the hydraulic systems. This enables researchers to develop methodologies of data-driven anomaly detection, aiming to detect and identify non-conforming behavior patterns during the operational process.

Data-driven anomaly detection methods can be broadly categorized into supervised, unsupervised, together with semi-supervised learning approaches. With the remarkable success of machine learning (ML), the advancement of computational power, and the accumulation of labeled datasets, supervised ML based anomaly detection algorithms have been proposed and successfully applied to various domains [2]–[6]. For instance, Akinyelu and Adewumi [7] developed a random forest based machine learning technique to classify phishing emails with 99.7% accuracy. Singh and Govindarasu [8] proposed the decision tree based anomaly detection methodology using differential features of voltage and current phasors to distinguish between normal tripping during power line faults and malicious tripping attacks on the physical relays in the remedial action scheme of smart grid. Hossain et.al [9] designed a Long Short-Term Memory (LSTM) based system for detecting and mitigating invalid in-vehicle Controller Area Network (CAN) messages and achieved high overall accuracy in detecting attacks.

When it comes to complex hydraulic systems, Helwig et al. [10] utilized linear discriminant analysis (LDA) to transfer significant condition-monitoring features to a lower dimensional discriminant space which was then used to classify anomaly conditions and grade of fault severity. Lei et al. [11] developed a novel fault diagnosis method by combining the advantages of Principal Component Analysis in dimensionality reduction and the advantages of the eXtreme Gradient Boosting (XGBoost) algorithm in classification. Compared with baselines, their proposed method can effectively identify valve faults in the hydraulic directional valve with high accuracy. Kim and Jeong [12] first applied data augmentation to increase data amount, then proposed a deep neural network model that integrates convolutional neural network (CNN), bidirectional long short-term memory network (BiLSTM), and attention mechanism for real-time hydraulic system condition monitoring which can also detect anomalous data. The model achieved better results than other deep learning models (e.g., pure CNN, LSTM).

The supervised approach relies on clear labeling of a large amount of training data, and their solutions usually define a clear decision boundary that separates normal and abnormal data. However, real-world data is often imperfect with labeled and unlabeled samples coexisting, leading to challenges. Therefore, unsupervised and semi-supervised approaches have emerged as alternative solutions. Unlike supervised ML, unsupervised ML does not require any additional information on the input data and aims to extract relevant characteristics from the data itself. Clustering [13], [14] and dictionary learning [15], [16] are examples of unsupervised methods applied for anomaly detection. Deep neural networks, for instance, deep belief networks and various types of autoencoders, have also been employed for unsupervised feature learning tasks [2], [17], [18].

In intricate systems, obtaining unsupervised models with high accuracy necessitates a solid understanding of the system, thereby making the process of training appropriate models more challenging. Given the trainer's limited knowledge of data distribution, a trial-and-error approach with constantly iterating through a large number of algorithms is therefore necessary. To address this issue, semi-supervised ML methods have been developed, which utilize unsupervised ML as a feature extraction mechanism to aid supervised ML when only a small number of labels are available. The algorithm is trained in combination with the trainer's knowledge of feature extraction, which is learned by applying it to anomaly detection via pilot testing. This approach is particularly useful in situations where condition monitoring devices collect a large amount of data, but only a small fraction of it is labeled and pertains to relevant to anomalous status. Moreover, in situations where the system is complex and there can be a large number of potential anomaly types that vary in nature and consequences, it is unrealistic to assume that data are available for every possible anomaly. In such cases, semi-supervised learning methods can also be highly beneficial.

However, there are rarely studies investigating semi-supervised learning for anomaly detection in hydraulic systems. To the knowledge of the authors, regarding the hydraulic system, only Yan et al. [19] developed a semi-supervised approach with unsupervised AEs using a multi-layer network structure for feature extraction and utilizing the distance-based similarity to form a health baseline trained from a large number of normal samples. The health baseline was adopted for anomaly detection for new observations. Furthermore, a comprehensive comparison is required to evaluate the performance of various semi-supervised ML algorithms regarding anomaly detection in hydraulic condition monitoring systems.

To fill the aforementioned research gaps, this study examined an open-sourced condition monitoring dataset of a complex hydraulic system [10], carried out a thorough data and feature analysis with effective feature engineering, and draw a systematical comparison regarding supervised ML applied for anomaly detection of this hydraulic system. Various supervised ML algorithms are incorporated and implemented, including traditional stand-alone models (e.g., one-class Support Vector Machines (SVM), Robust Covariance, Local Outlier Factor), ensemble models (e.g., Isolation Forest), and deep neural network based models (e.g., autoencoder, Hierarchical Extreme Learning Machine (HELM)). Extensive experiments show that the customized HELM model outperformed other semi-supervised algorithms obtaining the best performance with the highest accuracy (99.5%), the lowest false positive rate (0.015), and the best F1-score (0.985).

The remaining sections of this paper are organized as follows: Section II gives a brief introduction about the sensors and condition data used in this study. Section III carries out intensive data analysis and feature engineering. Section IV introduces the customized and implemented semi-supervised machine learning models and the utilized evaluation metrics for anomaly detection of hydraulic systems. Both traditional stand-alone models, ensemble models, and deep neural network based models are included. Section V presents the experiment design and model performance results. Section VI concludes the paper and proposes future research directions.

## II. SENSORS AND CONDITION DATA

The experimental study of this paper is based on a publicly available dataset produced by prior work [10]. It is important to understand the nature of this dataset and the described hydraulic system. Some essential properties are presented below, with further details deferred to the original source [10].

The test system is composed of two hydraulic circuits, depicted in Fig. 1. The colored labels highlight the key elements. The upper circuit is the primary working circuit, while the lower one is used for cooling and filtration purposes. The two circuits are connected through a common reservoir (oil tank). In the experiments conducted to generate the dataset, the proportional pressure-release valve V11 (located near the top-right of the schematic) was adjusted to generate various load levels and test conditions. The system is operated in cycles lasting 60 seconds each.

The sensors used in the dataset are summarized in Table I. The first column lists the abbreviated codes used to identify sensors in Fig. 1 and throughout the paper. The last three sensors in the table (CE, CP, and SE) are virtual sensors representing computed values, while the others correspond directly to physical sensors in the hydraulic circuit. The last column indicates the sampling rates of the sensors. The streaming data obtained from sampling these sensors is the primary input for data analysis.

The dataset also contains condition variables that describe the health status of some critical components in the hydraulic system, which are summarized in Table II. Similar to Table I, the abbreviations in the first column are used to identify the condition variables named in the second column throughout this paper. The third column lists the values to which these condition variables are set during the experimental work that generates the dataset, and the fourth column provides a maintenance-related interpretation of each value. The values of these condition variables are used as the ground truth for the anomaly detection task. The dataset presents the above sensor and condition data in their original form, without feature extraction or similar processing. All data, for both sensor and condition variables, from one 60-second cycle is grouped together in one conceptual row. The dataset contains 2205 such rows, i.e., 2205 minutes and 2205 samples.

The sensor files contain a temporally ordered sequence of values for each sensor, with sensors sampled at 100 Hz having

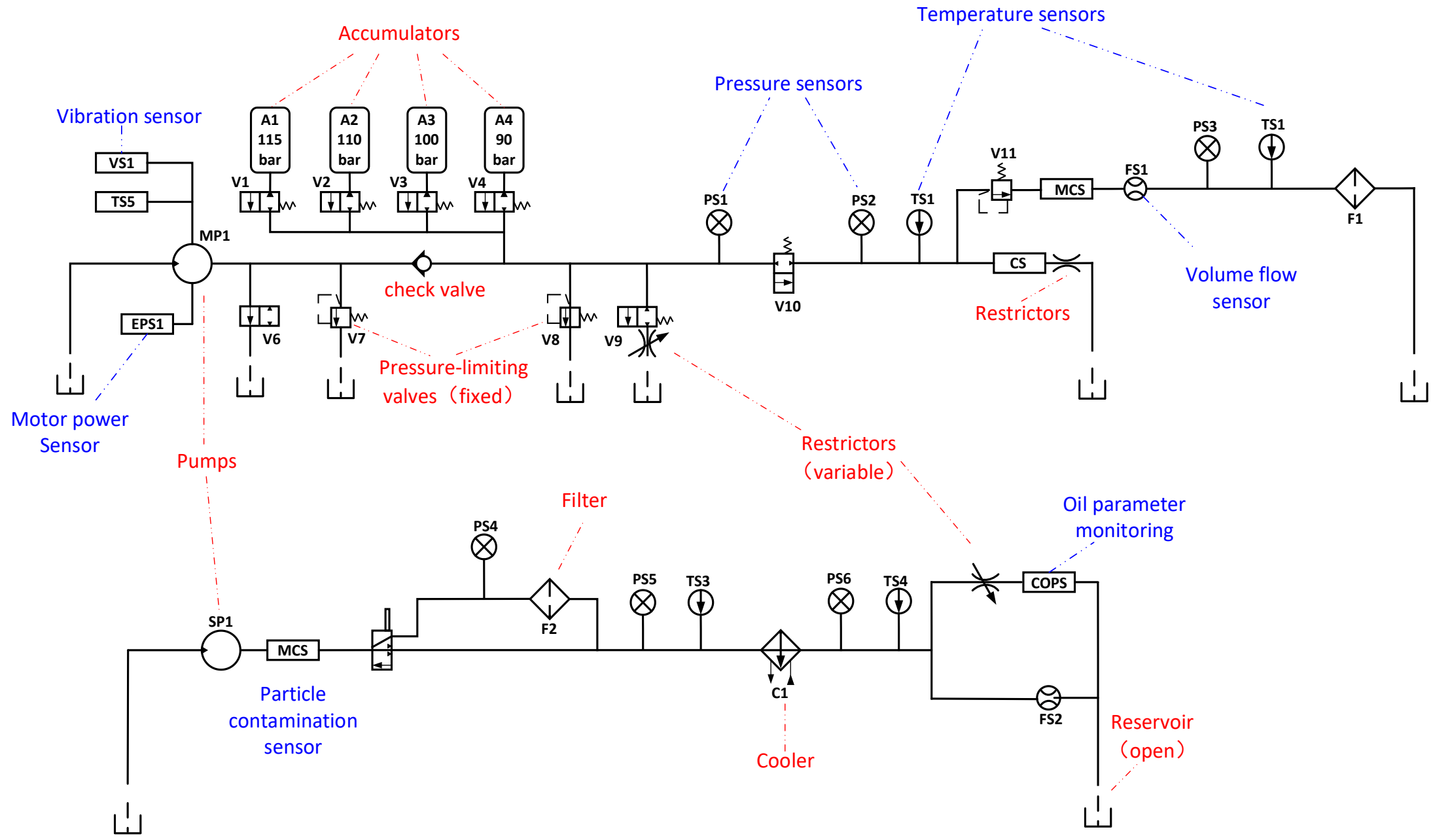

Fig. 1. A schematic diagram of the hydraulic test rig underlying the condition monitoring hydraulic system dataset [10]. Sensors are labeled in blue and other elements are labeled in red.

TABLE I. SENSORS USED BY THE HYDRAULIC TEST RIG IN FIG. 1

| Sensor | Physical quantity | Unit | Rate (Hz) |
|---|---|---|---|
| PS1--PS6 | Pressure | bar | 100 |
| EPS1 | Motor power | W | 100 |
| FS1,FS2 | Volume flow | l/min | 10 |
| TS1--TS4 | Temperature | C | 1 |
| VS1 | Vibration | mm/s | 1 |
| CE | Cooling efficiency (virtual) | % | 1 |
| CP | Cooling power (virtual) | kW | 1 |
| SE | Efficiency factor | % | 1 |

TABLE II. CONDITION VARIABLES IN THE HYDRAULIC TEST RIG IN FIG. 1

| Abbr. | Variable (unit) | Value | Interpretation |
|---|---|---|---|
| cool | Cooler condition (%) | 3<br>20<br>100 | close to total failure<br>reduced efficiency<br>full efficiency |
| valv | Valv condition (%) | 100<br>90<br>80<br>73 | optional switching<br>small lag<br>severe lag<br>close to total failure |
| **pump** | **Internal pump leakage (code)** | **0**<br>**1**<br>**2** | **no leakage**<br>**weak leakage**<br>**severely leakage** |
| hydr | Hydraulic accumulator (code) | 130<br>115<br>100<br>90 | optional pressure<br>slightly low pressure<br>severely low pressure<br>close to total failure |
| stab | Stable flag (code) | 0<br>1 | conditions stale<br>may be unstable |

6000 values per row, sensors sampled at 10 Hz having 600 values per row, and sensors sampled at 1 Hz having 60 values.

The condition variable file lists a single value for each of the five condition variables in each row.

In this study, the "internal pump leakage" status is chosen with "no leakage" as normal and "weak leakage" together with "*severely leakage*" as anomalies. Given the sensor readings for each instance (sample), the task is to predict the leakage status.

## III. DATA ANALYSIS AND FEATURE ENGINEERING

The process of feature engineering involves creating more suitable and informative features for model training by combining or integrating existing features in new ways. While it is important to make use of all available information, features that are particularly relevant for the leakage status detection task can be emphasized to produce a more effective model.

When examing the nature and volume of data in each instance, it is found that each instance has a large number of attributes (43680 attributes/instance) due to the combination of sensor readings taken at different sampling rates, i.e., [(100 Hz $\times$ 7 sensors) + (10 Hz $\times$ 2 sensors) + (1 Hz $\times$ 8 sensors)] $\times$ 60 s/instance. Conventional machine learning methods are unsuitable for handling data with such a large number of attributes. Using the instances from the dataset directly as inputs ignores the important fact that most of the attributes form a time series of sensor values and within each second the values usually fluctuate around their means. A well-established strategy is to compute a collection of series-oriented statistics and transforms, such as skewness, instead of using the original attributes. Following this strategy, the time series for each of the 17 sensors in each instance is replaced by four standard statistics, i.e., *mean*, *variance*, *skewness*, and *kurtosis*. This procedure yields instances with 17 $\times$ 4 = 68 attributes. Histogram visualizations for selected features are shown in Fig. 2. It is observed that the values of these features distribute in several clusters. Fig. 3 shows the correlation matrix between the features. Some of the features present higher correlations than others. Furthermore, Fig. 4 illustrates the dimensionality reduction results using the t-distributed stochastic neighbor embedding (t-SNE) method. It is observed that there are some clusters regarding *leakage* and

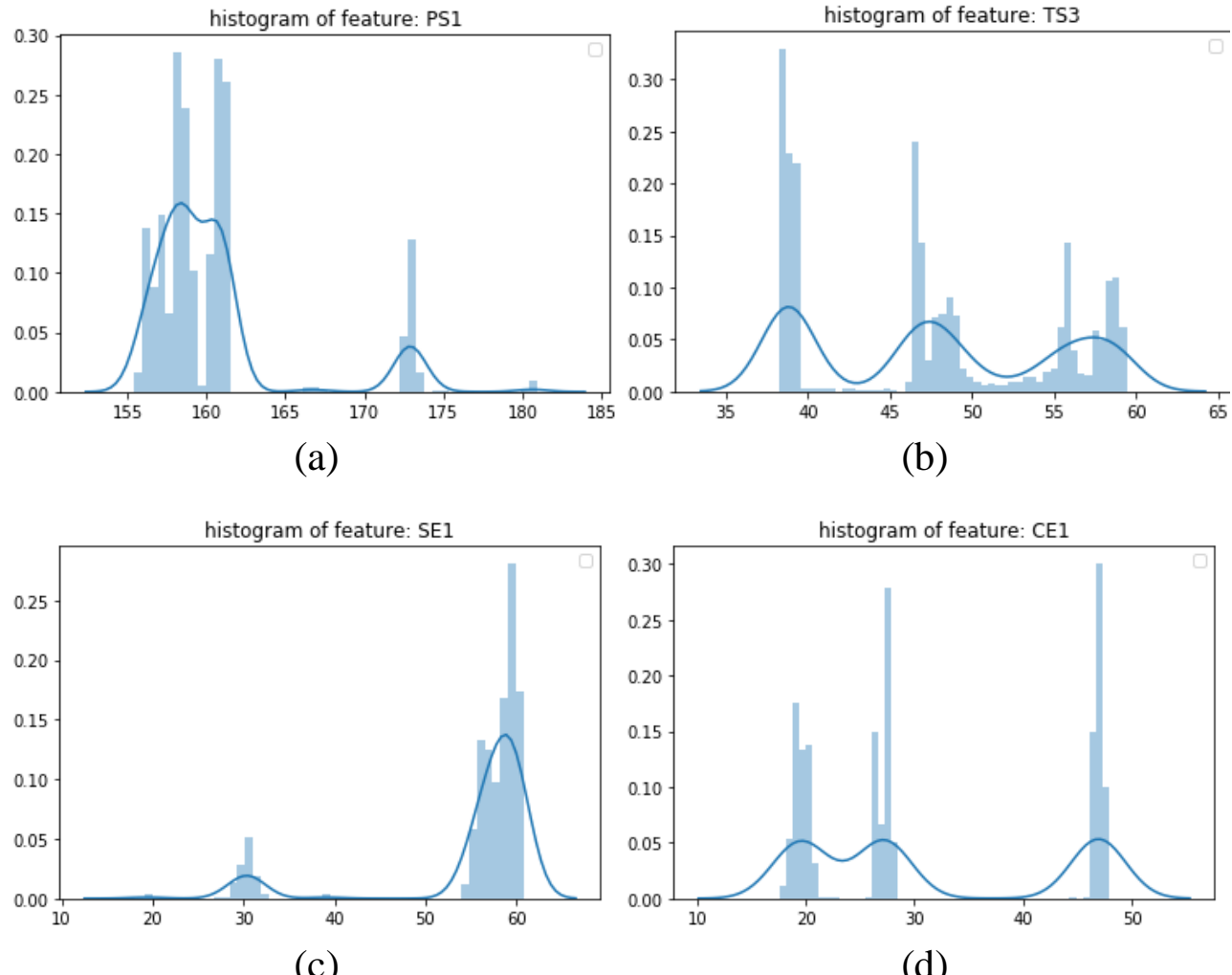

Fig. 2. Histogram visualizations for selected features: (a) PS1, (b) TS3, (c) SE1, and (d) CE1.

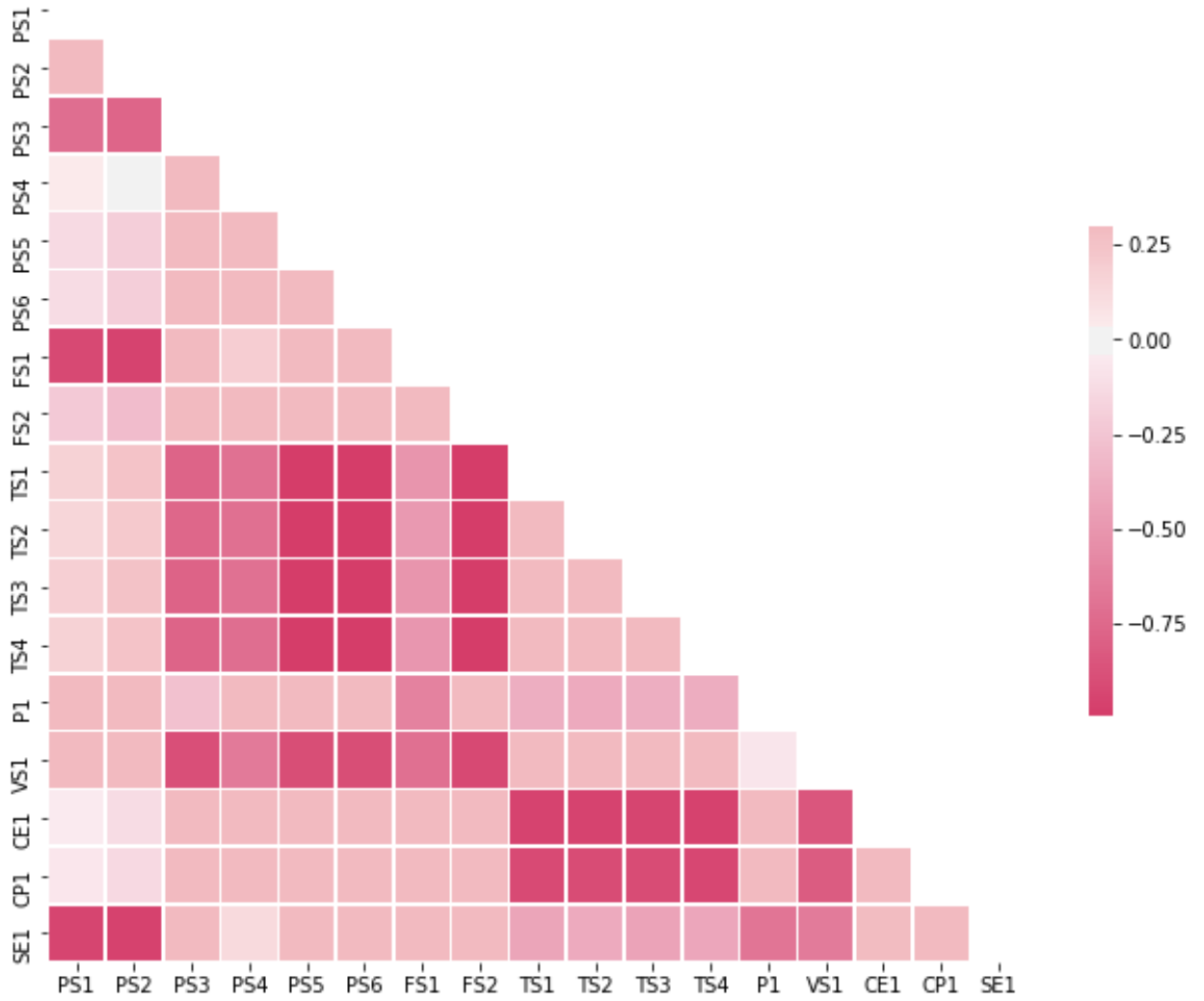

Fig. 3. Correlation matrix between the features.

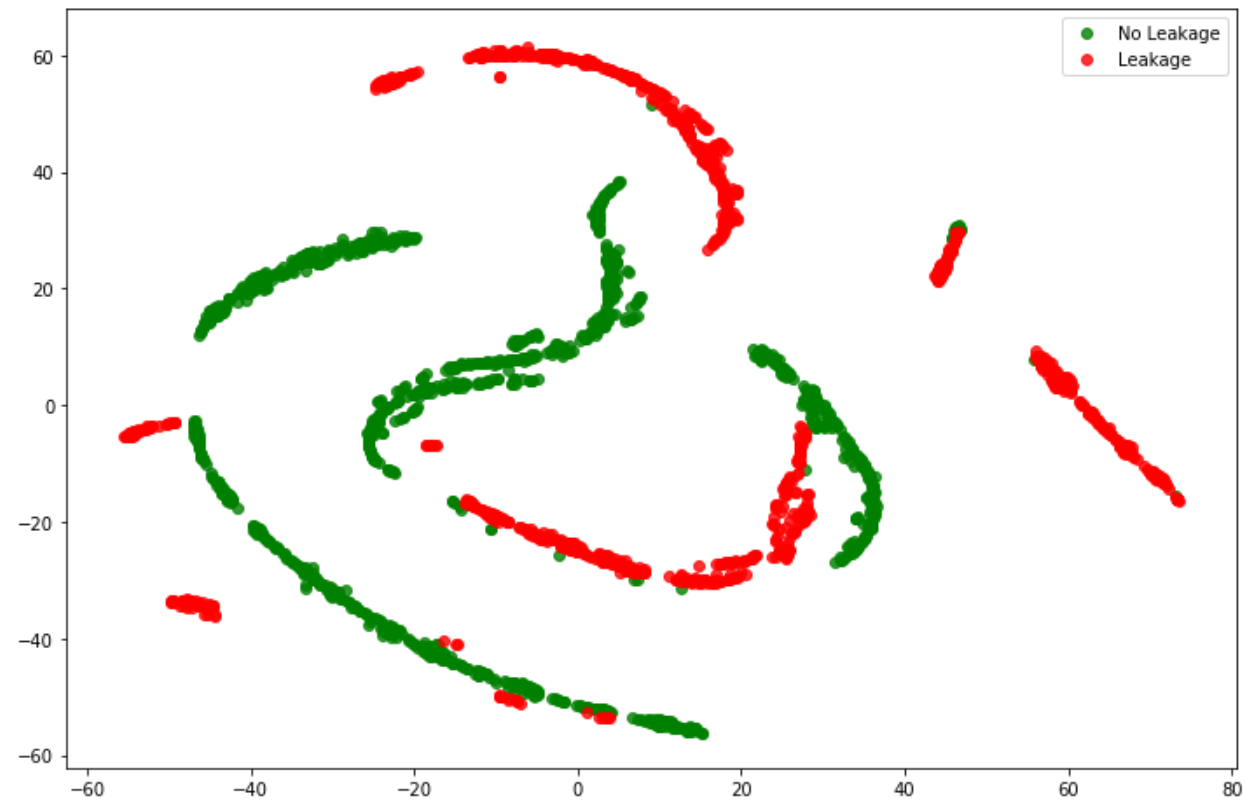

Fig. 4. Dimensionality reduction visualizations using t-SNE.

*no leakage* status, however, no clear cluster decision boundary can be set.

After feature engineering, various semi-supervised ML algorithms were implemented and employed to build anomaly detection models.

## IV. Semi-Supervised Machine Learning for Anomaly Detection of Hydraulic System

### A. *Stand-alone Semi-Supervised Machine Learning Models*

Regarding stand-alone semi-supervised machine learning methods, this study selected Robust Covariance (RC), Local Outlier Factor (LOF), and one-class SVM as the baselines for anomaly detection in the hydraulic condition monitoring system.

Robust covariance (RC) is a variance-based robust anomaly detection model which assumes that all normal samples obey Gaussian distribution and uses the Mahalanobis distance to derive the outliers.

The Local Outlier Factor (LOF) considers the density of data points in the actual distribution and utilizes local density as a key factor to detect outliers. The locality is calculated by the k-nearest neighbors (KNN) algorithm. LOF algorithm determines the outliers in two dimensions, one is the small reachable density of the target sample points, and the other is the large reachable density of all the k-nearest neighbors of the target samples.

One-class Support Vector Machine (one-class SVM) works on the basic idea of minimizing a hypersphere of the single class of examples in the training data and considers all the other samples outside the hypersphere to be outliers/anomalies or out of training data distribution. The training and model fitting of one-class SVM is to minimize the radius of a hypersphere. However, as the standard one-class SVM is very restrictive to outliers in the training set, to make it more flexible to tolerate outliers, a penalty factor is usually given.

### B. *Semi-Supervised Ensemble Learning Models*

Isolation Forest (IF) is chosen as the representative of semi-supervised ensemble learning models. IF is more relaxed in terms of data features, it splits the data space using lines that are orthogonal to the origin and assigns higher scores to the data points that need fewer splits to be isolated. Because IF separates outliers by successively cutting subspaces, its performance becomes more stable when increasing the number of trees. Compared with other traditional algorithms, such as LOF, IF has no assumptions on the distribution of the data set, which means it is more robust. However, IF has limitations when dealing with high-dimensional data and a high number of local outliers.

### C. *Deep Learning Based Semi-Supervised Models*

In this study, deep autoencoder (DAE) and Hierarchical Extreme Learning Machine (HELM) based semi-supervised models were customized and implemented for the anomaly detection task in the hydraulic system.

#### ***Deep Autoencoder Based Semi-supervised Learning***

The DAE-based semi-supervised learning model is constructed by two symmetrical feedforward multilayer neural networks, i.e., the *encoder* and the *decoder*. The input data samples are first fed to the encoder part for feature extraction, through which the compressed feature vector is obtained. To capture the most critical features representing the input data, fewer intermediate hidden layer nodes are used to obtain the compressed feature vector which removes redundant information and keeps the most significant ones. Then the decoder decodes the compressed feature vector into the original dimension reconstructing the input.

In the training phase, normal data samples were input to train the autoencoder's weights by minimizing the reconstruction loss as the objective. Backpropagation is adopted as the updating method. In the validation phase, a validation dataset was adopted to fine-tune a good threshold of the reconstruction loss. The validation dataset contains normal data samples which are unseen in the training phase and a few labeled anomaly data samples. Finally in the application phase, for new coming data samples, after feeding them through the trained autoencoder, the reconstruction losses are calculated and then compared with the fine-tuned threshold for anomaly detection. More details about DAE-based semi-supervised learning can be found in [18].

### *HELM Based Semi-supervised Learning*

This study customized the Hierarchical Extreme Learning Machine (HELM) model and implemented it for the anomaly detection task. HELM is based on the hierarchical stacking of Extreme Learning Machines (ELMs).

Assume that there are $N$ arbitrary samples $(\boldsymbol{X}, \boldsymbol{T})$, $\boldsymbol{X} = [\boldsymbol{x}_1, \boldsymbol{x}_2, \cdots, \boldsymbol{x}_N]^T, \boldsymbol{T} = [\boldsymbol{t}_1, \boldsymbol{t}_2, \cdots, \boldsymbol{t}_N]^T$, where $\boldsymbol{x}_i$ is the feature and $\boldsymbol{t}_i$ stands for the desired output target.

For a single hidden layer ELM neural network with $L$ hidden nodes, the output $\boldsymbol{O} = [\boldsymbol{o}_1, \boldsymbol{o}_2, \cdots, \boldsymbol{o}_N]^T$ can be expressed as

$$\sum_{i=1}^{L} \beta_i g\left(\boldsymbol{W}_{\boldsymbol{i}} \cdot \boldsymbol{x}_{\boldsymbol{j}} + b_j\right) = \boldsymbol{o}_j,\ j = 1, 2, \cdots, N \tag{1}$$

where g(x) is the activation function, $\beta_{\mathrm{i}}$ is the output weight, $W_i$ is the input weight and $b_j$ is the $j$th bias of the first hidden layer.

Ideally, there should be

$$\sum_{j=1}^{N} \left\| \boldsymbol{o}_j - \boldsymbol{t}_{\boldsymbol{j}} \right\| = 0 \tag{2}$$

that is, there exists $\boldsymbol{\beta}_i$, $\boldsymbol{W}_{\mathrm{i}}$ and $b_i$ such that

$$\sum_{i=1}^{L} \boldsymbol{\beta}_i g\left(\boldsymbol{W}_i \cdot \boldsymbol{x}_j + b_j\right) = \boldsymbol{t}_j, j = 1, 2, \cdots, N \tag{3}$$

which can be represented by matrixes as

$$\boldsymbol{H\beta} = \boldsymbol{T} \tag{4}$$

where $\boldsymbol{H}$ is the output of the hidden layer node, $\boldsymbol{\beta}$ is the output weight, and $\boldsymbol{T}$ is the desired output.

$$H(\boldsymbol{W}_1, \cdots, \boldsymbol{W}_L, b_1, \cdots, b_L, \boldsymbol{x}_1, \cdots, \boldsymbol{x}_N) = \begin{bmatrix} g(\boldsymbol{W}_1 \cdot \boldsymbol{x}_1 + b_1) & \cdots & g(\boldsymbol{W}_L \cdot \boldsymbol{x}_1 + b_L) \\ \vdots & \cdots & \vdots \\ g(\boldsymbol{W}_1 \cdot \boldsymbol{x}_N + b_1) & \cdots & g(\boldsymbol{W}_L \cdot \boldsymbol{x}_N + b_L) \end{bmatrix}_{N \times L} \tag{5}$$

To train the single hidden layer ELM neural network is equivalent to obtaining $\widehat{\boldsymbol{\beta}}$ such that

$$\left\| \boldsymbol{H}\widehat{\boldsymbol{\beta}} - \boldsymbol{T} \right\| = \min_{\boldsymbol{\beta}} \| \boldsymbol{H\beta} - \boldsymbol{T} \| \tag{6}$$

When choosing the mean square error (MSE) as the measure, (6) is equivalent to minimizing the following loss function

$$Loss = \sum_{j=1}^{N} \left( \sum_{i=1}^{L} \boldsymbol{\beta}_i g(\boldsymbol{W}_i \cdot \boldsymbol{x}_j + b_i) - \boldsymbol{t}_j \right)^2 \tag{7}$$

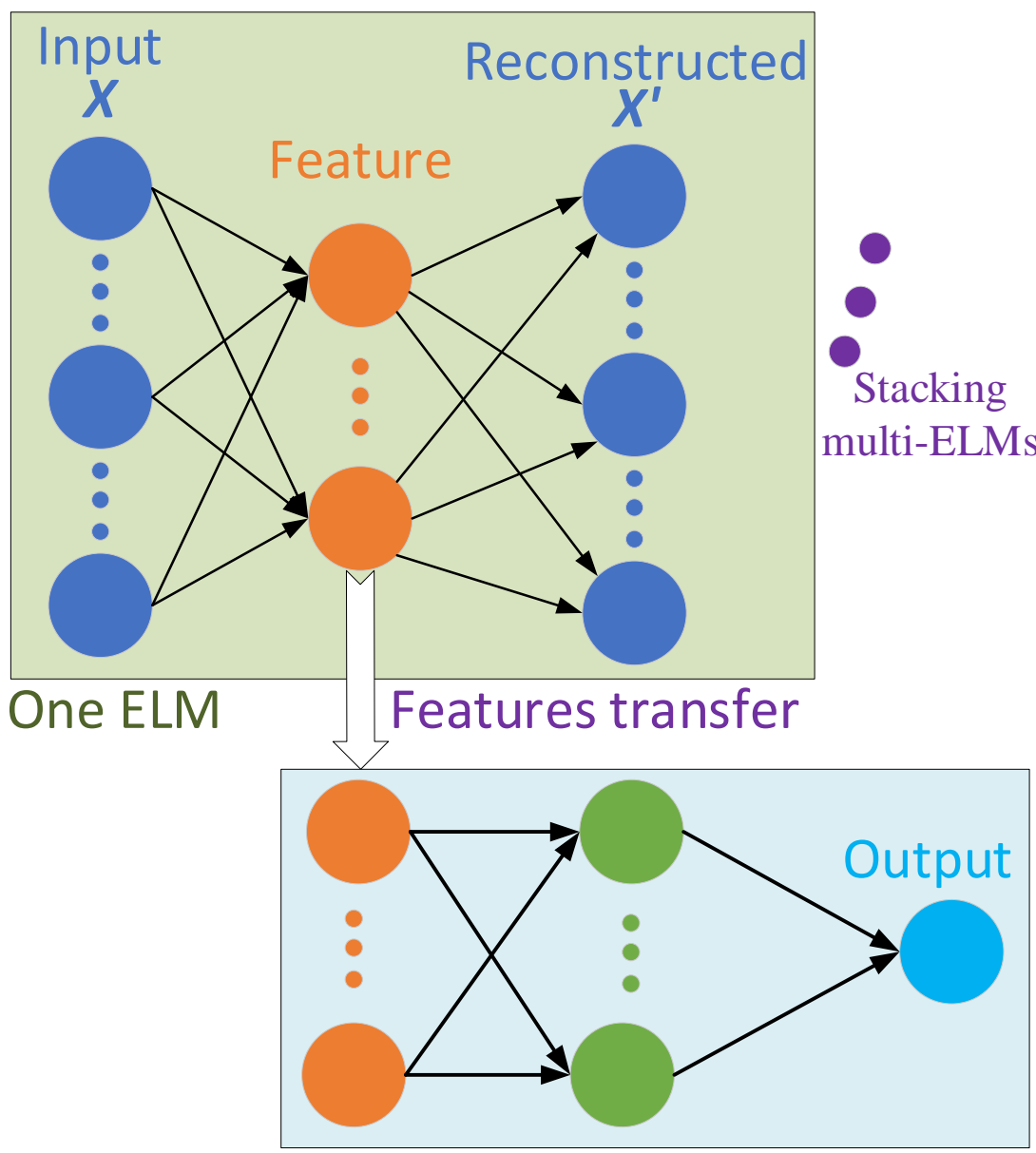

Fig. 5. Framework of HELM-based semi-supervised method.

The ELM allows the weights $\boldsymbol{\beta}$ and the deviation between the hidden layer and the inputs to have random values that can be sampled from any distribution. Thus, the learning step consists only in finding the optimal weight $\boldsymbol{\beta}$ between the hidden layer and the output. The drawback of the pure ELM is that its shallow architecture cannot effectively handle data contents, even with a large number of hidden nodes. HELM which hierarchically stacks multi-layers of ELM is one of the most successful attempts to create a deeper structure based on the ELM principles. Therefore, HELM is introduced.

In this study, hierarchical ELM layers were first trained using only normal data without any anomalies. By minimizing the reconstruction loss, the ELMs can capture the most critical features representing the input data. Then the captured features are transferred to the one-class classifier which is further trained to obtain a threshold using a validation dataset that is totally unseen during the training process. The validation dataset also only contains normal data samples. Usually, a good threshold $Thrd$ can be expressed by

$$Thrd = \gamma \cdot \mathrm{percentile}_p(|1 - \boldsymbol{Y}^{\mathrm{valid}}|) \tag{8}$$

where $\boldsymbol{Y}^{\mathrm{valid}}$ is the output of the one-class classifier, $percentile_p$ is a function of the $p$th percentile with hyperparameters $p$ and $\gamma \geq 0$.

Finally in the application testing phase, for new coming data samples, after feeding them through the trained HELM, the outputs of the one-class classifier $\boldsymbol{Y}^{\mathrm{test}}$ will be compared with the $Thrd$, and the label can be obtained by

$$Label_{\boldsymbol{Y}^{\mathrm{test}}} = \mathrm{sgn}\ (Thrd - |1 - \boldsymbol{Y}^{\mathrm{test}}|) \tag{9}$$

The framework of HELM-based semi-supervised method is illustrated in Fig. 5.

### *D. Evaluation Metrics*

Various metrics are used to evaluate the overall performance of the selected model. Four basic terms, i.e.,

True-positive (TP) which represents the number of correctly detected anomalies, True-negative (TN) which represents the number of correctly detected normals, False-positive (FP) which represents the number of incorrectly detected anomalies, and False-negative (FN) which represents the number of incorrectly detected normals, are first obtained. Then, based on the four terms, accuracy, precision, recall (true positive rate), and false positive rate (FPR) were calculated.

Accuracy (ACC) is the percentage of correctly predicted samples in the total sample, whose mathematical expression can be defined as follows:

$$Accuracy = \frac{\text{TP+TN}}{\text{TP+TN+FP+FN}} \tag{10}$$

Precision is a metric that measures the proportion of correctly predicted positive samples out of all predicted positive samples. It indicates the reliability of the positive predictions. The mathematical expression for precision is:

$$Precision = \frac{\text{TP}}{\text{TP+FP}} \tag{11}$$

Recall ratio is the percentage of positive observations correctly predicted in the actual category. It is equivalent to the true positive rate (TPR) and can be calculated by

$$Recall = \text{TPR} = \frac{\text{TP}}{\text{TP+FN}} \tag{12}$$

False positive rate (FPR) is calculated as the ratio between the number of negative samples wrongly categorized as positive (false positives, FP) and the total number of actual negative samples, i.e.,

$$\text{FPR} = \frac{\text{FP}}{\text{FP+TN}} \tag{13}$$

Finally, the F1-score provides an overall view of recall and precision (weighted average). F1-score ranges from 0.0 to 1.0, with 1.0 indicating perfect precision and recall.

$$\text{F1} - \text{score} = 2 \times \frac{Precision \times Recall}{Precision + Recall} \tag{14}$$

## V. EXPERIMENT AND RESULTS

To test the selected semi-supervised models, the dataset was split into the train, valid, and test sets, with train and valid sets only containing normal data samples, and the majority of the test set being anomalies. Table III synthesizes the quantitative performance comparison results of the selected semi-supervised ML models. As shown in the table, HELM provides the best performance obtaining the highest accuracy (99.5%), the lowest false positive rate (0.015), and the best F1-score (0.985) beating other semi-supervised methods. It is observed that the Robust Covariance performs significantly worse: there are the lowest ACC, F1_score, and TPR, together with the highest FPR. Isolation forest obtains the best TPR at 1.000 which means it correctly detects all anomalies. The deep autoencoder based model does not perform better than the traditional ML models, (i.e., Local Outlier Factor, one-class SVM, and Isolation Forest), which further demonstrates HELM's superiorities of feature learning in this task since they are both deep learning based approaches. To better visualize the results, Fig. 6 provides the confusion matrix visualizations for the models from which one can identify the correctly classified number of normal and anomaly instances.

Furthermore, Fig. 7 and Fig. 8 visualize the distribution of $\frac{|1-\boldsymbol{Y}^{test}|}{Thrd}$ obtained from the HELM based semi-supervised learning method. It can be seen that the anomaly data instances (i.e., leakage ones) will obtain higher values of $\frac{|1-\boldsymbol{Y}^{test}|}{Thrd}$ with the chosen threshold $Thrd$ can perfectly separate normal and anomalous samples. This also verifies the effectiveness of the customized HELM method. However, using the developedHELM is impossible to detect the anomalous type (i.e., weak leakage or severe leakage) since the *weak leakage* samples and the *severe leakage* samples demonstrate similar distributions of $\frac{|1-\boldsymbol{Y}^{test}|}{Thrd}$ as shown in Fig. 8. Semi-supervised learning methods that can detect the potential anomalous types need further exploration in future studies.

TABLE III. MODEL PERFORMANCE COMPARISON

| Model | ACC | TPR | FPR | F1_Score |
| --- | --- | --- | --- | --- |
| Robust Covariance | 0.857 | 0.859 | 0.148 | 0.665 |
| Local Outlier Factor | 0.973 | 0.986 | 0.092 | 0.918 |
| One-class SVM | 0.975 | 0.995 | 0.122 | 0.922 |
| Isolation Forest | 0.989 | 1.000 | 0.066 | 0.966 |
| Deep Autoencoder | 0.985 | 0.975 | 0.128 | 0.872 |
| HELM | **0.995** | 0.997 | **0.015** | **0.985** |

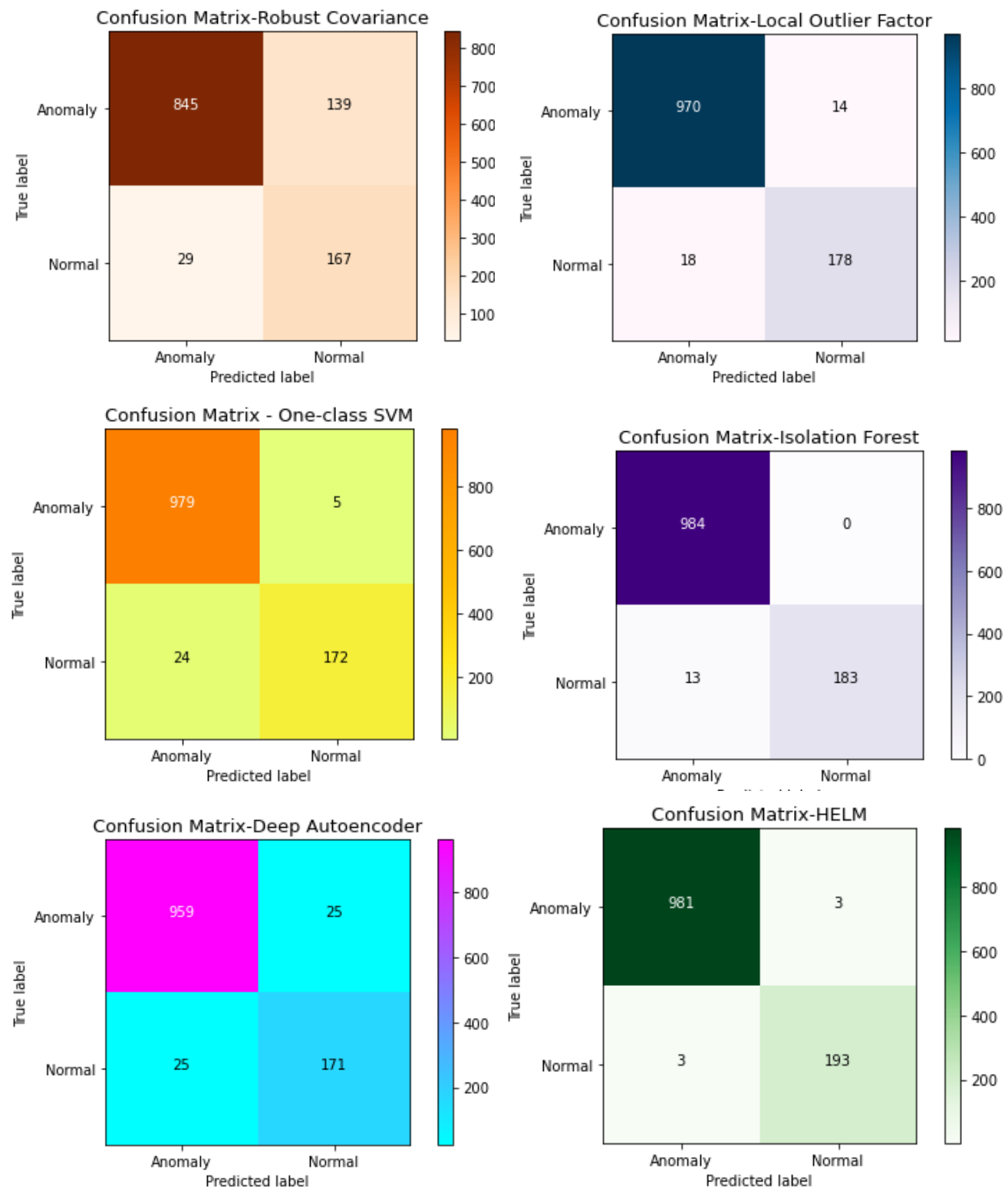

Fig. 6. The confusion matrixes of tested semi-supervised models.

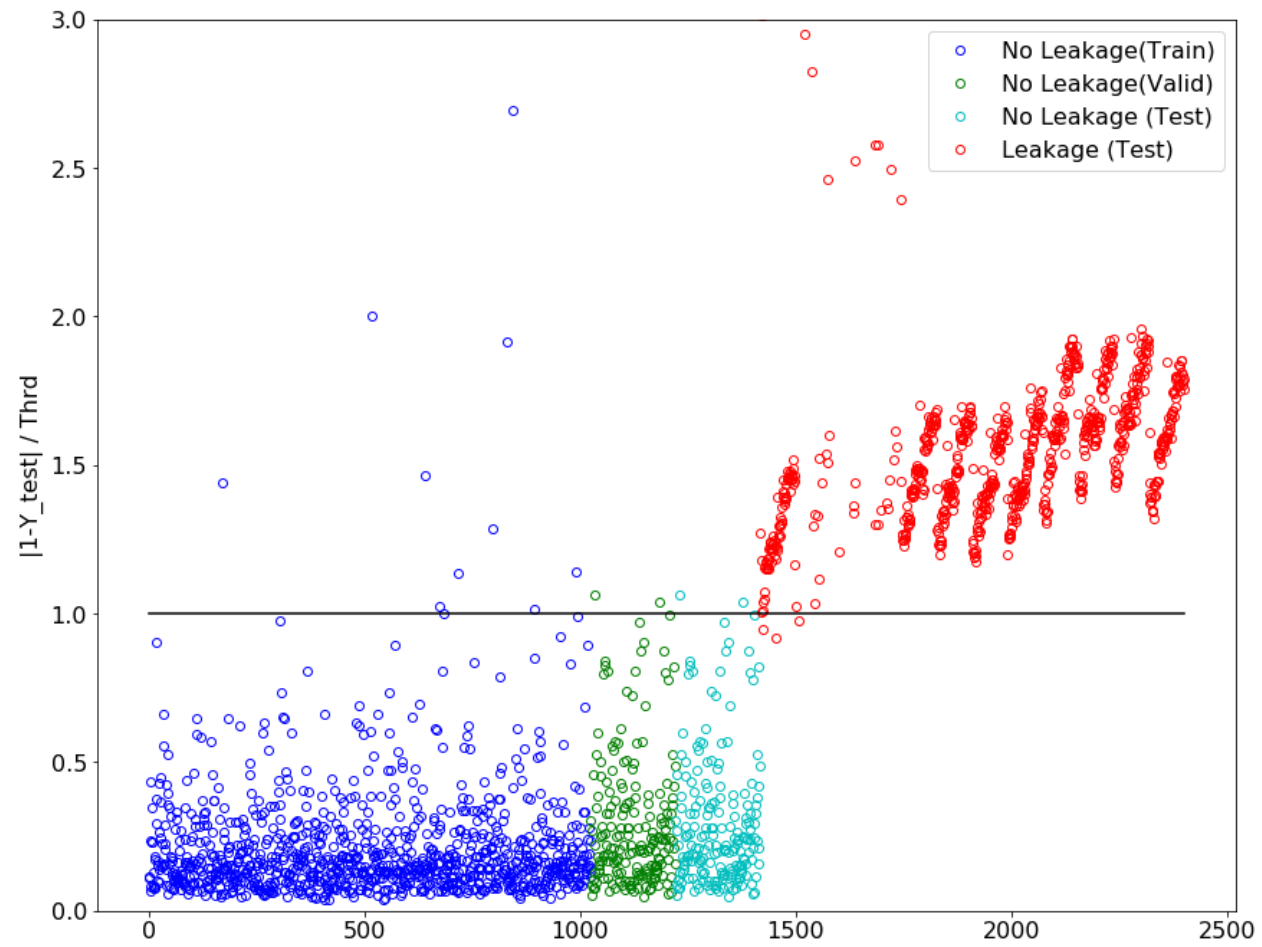

Fig. 7. The visualization for $|1 - \boldsymbol{Y}^{test}|/Thrd$ of the HELM method.

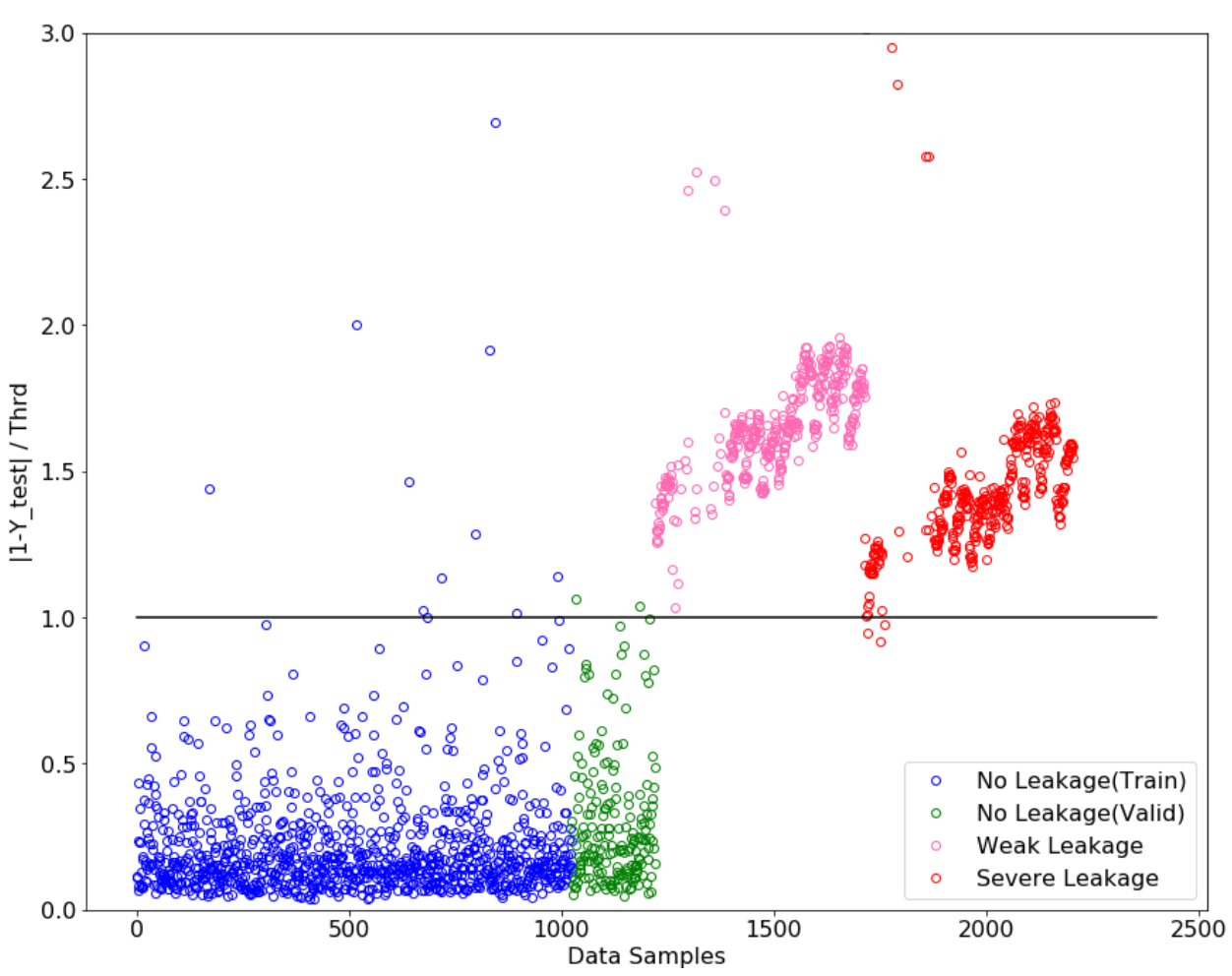

Fig. 8. The visualization for $|1-\boldsymbol{Y}^{test}|/Thrd$ of the HELM method regarding potential fault type detection.

## VI. CONCLUSION AND DISCUSSION

To enhance the condition monitoring of hydraulic systems regarding potential anomaly detection and promote research on semi-supervised learning methods, this study examined an open-sourced condition monitoring hydraulic system dataset carrying out thorough feature engineering and data analysis. With the extracted valuable and effective features, this study implemented, tested, and compared various semi-supervised machine learning models. As few studies have employed semi-supervised learning methods for hydraulic systems, this research fills the research gap and comprehensively compares traditional stand-alone semi-supervised learning models (e.g., one-class SVM, Robust Covariance), ensemble learning based models (e.g., Isolation Forest), and deep neural network based models (e.g., deep autoencoder, HELM). Furthermore, the customized HELM model obtained state-of-the-art performance with the best accuracy, false positive rate, and F1-score, beating all other semi-supervised methods examined. From the visualization of the indicator with regard to the tuned threshold for both normal and anomaly samples, the effectiveness of the HELM method can be verified, although further explorations need to be investigated.

Visualization also demonstrates that the developed HELM model can not be employed for identifying the potential anomalous types. It is suggested to further investigate semi-supervised methods that can detect the potential fault types in future studies.